\documentclass[runningheads]{llncs}

\usepackage{eccv}

\PassOptionsToPackage{table}{xcolor}
\usepackage{booktabs}
\usepackage{multirow}
\usepackage{wrapfig}

\definecolor{bestColor}{RGB}{200,215,235}   
\definecolor{secondColor}{RGB}{215,220,225} 
\definecolor{worstColor}{RGB}{226,126,126} 
\newcommand{\besttxt}[1]{\colorbox{bestColor}{\textbf{#1}}}
\newcommand{\secondtxt}[1]{\colorbox{secondColor}{\textbf{#1}}}
\newcommand{\worsttxt}[1]{
  \ifnum\value{table}=2 #1
  \else\ifnum\value{table}=3 #1
  \else\colorbox{worstColor}{\textbf{#1}}
  \fi\fi
}

\newcommand{\best}[1]{\cellcolor{bestColor}\textbf{#1}}
\newcommand{\second}[1]{\cellcolor{secondColor}\underline{#1}}
\newcommand{\worst}[1]{
  \ifnum\value{table}=2 #1
  \else\ifnum\value{table}=3 #1
  \else\cellcolor{worstColor}{#1}
  \fi\fi
}

\newcommand\blfootnote[1]{
  \begingroup
  \renewcommand\thefootnote{}\footnote{#1}
  \addtocounter{footnote}{-1}
  \endgroup
}

\usepackage{booktabs}      
\usepackage{multirow}
\usepackage{graphicx}
\usepackage{colortbl}      

\usepackage{eccvabbrv}

\usepackage{booktabs}
\usepackage{amsmath} 
\usepackage{amssymb} 
\usepackage{graphicx} 
\usepackage{makecell}
\usepackage{multirow}
\usepackage{pifont}
\usepackage{adjustbox}
\usepackage{enumitem}
\usepackage{tikz}
\usepackage{tabularx}
\usepackage{algorithm}
\usepackage[english]{babel}
\usepackage{algorithmic}
\usepackage{bbold}
\usepackage[accsupp]{axessibility}  

 \usepackage{hyperref}

\usepackage{orcidlink}

\newcolumntype{L}[1]{>{\raggedright\arraybackslash}p{#1}}
\newcolumntype{C}[1]{>{\centering\arraybackslash}p{#1}}
\newcolumntype{R}[1]{>{\raggedleft\arraybackslash}p{#1}}

\newcommand{\mypara}[1]{\noindent\textbf{#1}}

\begin{document}

\title{EgoGrasp: World-Space Hand-Object Interaction Estimation from Egocentric Videos} 

\titlerunning{EgoGrasp}

\author{
Hongming Fu\inst{1} \and
Wenjia Wang\inst{2\dagger} \and 
Xiaozhen Qiao\inst{3} \and 
Rolandos Alexandros Potamias\inst{4} \and 
Taku Komura\inst{2} \and 
Shuo Yang\inst{5} \and 
Zheng Liu\inst{6} \and 
Bo Zhao\inst{1\ddagger}
}

\authorrunning{H. Fu, W. Wang et al.}

\institute{\scriptsize
\mbox{Shanghai Jiao Tong University \and
The University of Hong Kong}
\mbox{\and University of Science and Technology of China \and
Imperial College London }
\mbox{\and  Harbin Institute of Technology (Shenzhen) \and
Beijing Academy of Artificial Intelligence} \\
\email{\{fuhongming,bo.zhao\}@sjtu.edu.cn}
\email{wwj2022@connect.hku.hk} \\
\url{https://mint-sjtu.github.io/EgoGrasp.io}
}

\maketitle

\blfootnote{$\dagger$: Project lead, $\ddagger$: Corresponding author}
\vspace{-25pt}

\begin{abstract}

We propose \textbf{EgoGrasp}, the first method to reconstruct world-space hand-object interactions (W-HOI) from dynamic egoview videos, supporting open-vocabulary objects. Accurate W-HOI reconstruction is critical for embodied intelligence yet remains challenging. Existing HOI methods are largely restricted to local camera coordinates or single frames, failing to capture global temporal dynamics. While some recent approaches attempt world-space hand estimation, they overlook object poses and HOI constraints. Moreover, previous HOI estimation methods either fail to handle open-set categories due to their reliance on object templates or employ differentiable rendering that requires per-instance optimization, resulting in prohibitive computational costs. Finally, frequent occlusions in egocentric videos severely degrade performance. To overcome these challenges, we propose a multi-stage framework: (i) a robust pre-processing pipeline leveraging vision foundation models for initial 3D scene, hand and object reconstruction; (ii) a body-guided diffusion model that incorporates explicit egocentric body priors for hand pose estimation; and (iii) an HOI-prior-informed diffusion model for hand-aware 6DoF pose infilling, ensuring physically plausible and temporally consistent W-HOI estimation. We experimentally demonstrate that EgoGrasp can achieve state-of-the-art performance in W-HOI reconstruction, handling multiple and open vocabulary objects robustly. 
\end{abstract}

\begin{figure}[!ht]
    \centering
    \vspace{-27pt}
    \includegraphics[height=6cm, width=0.99\textwidth, keepaspectratio]{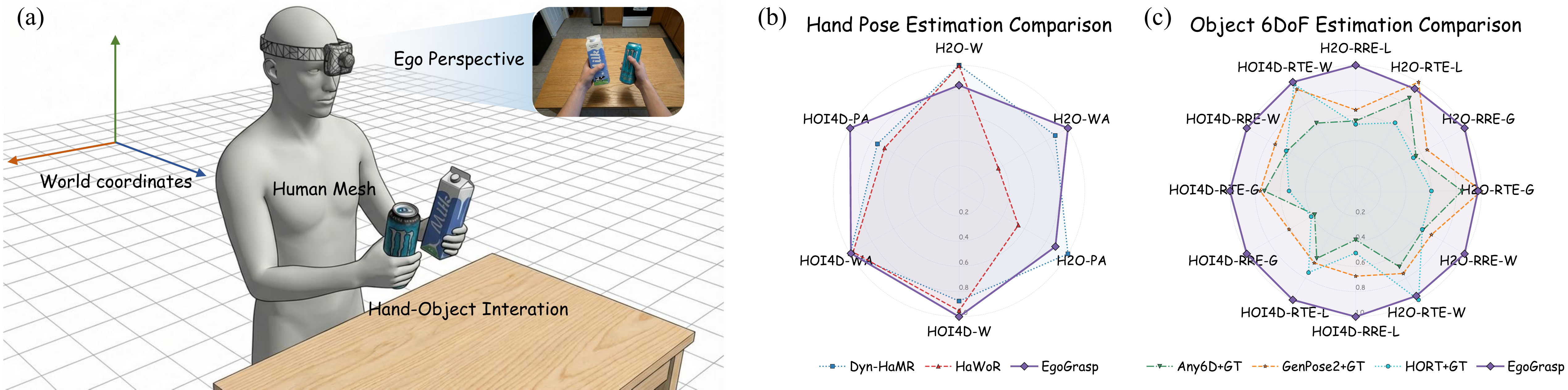}
    \vspace{-5pt}
    \caption{(a) shows that EgoGrasp reconstructs world-space hand-object interactions from egocentric videos with dynamic cameras. (b) and (c) highlight the superior performance of EgoGrasp in estimating hand-object interactions in world space.}
    \vspace{-10pt}
    \label{fig:teaser}
\end{figure}

\section{Introduction}

Understanding hand-object interaction from egocentric videos is a cornerstone of computer vision and embodied intelligence. Reconstructing accurate world-space HOI poses is crucial for analyzing human manipulation behavior and enabling downstream applications in embodied AI, robotics, and virtual/augmented reality~\cite{luo2025being, yang2025egovla, li2025maniptrans}. While egocentric videos provide unique first-person cues, they are typically recorded by dynamic cameras in unconstrained environments, where relying solely on per-frame camera-space geometry is insufficient. To fully interpret human actions, we argue that recovering temporally coherent trajectories in world space is essential to address three fundamental challenges: 
1) Incorporating world-space is essential for stabilizing interaction dynamics. In egocentric scenarios, high-frequency camera jitter is entangled with object movement in local views. By anchoring the motion to the static world, we effectively filter out this noise, ensuring the reconstruction reflects firm physical grasps rather than unrealistic contact sliding.
2) Furthermore, static background cues could resolve semantic ambiguities. E.g., a user leaning towards an object is otherwise indistinguishable from lifting an object in the local frame. Global context clarifies the true source of motion, ensuring accurate trajectory reconstruction~\cite{wang2026embodmocap}.
3) World space helps ground the object within the static environment, enforcing geometric constraints that prevent objects from physically violating the scene, such as floating above or penetrating table surfaces during placement.

Despite progress in 3D hand and 6DoF object estimation, existing methods fail to predict world-space HOI in egocentric videos. Most approaches operate at the image or short-sequence level, predicting results in camera coordinates~\cite{liu2021semi,cao2021reconstructing,ye2023diffusion,ye2024g,wen2024foundationpose,lee2025any6d}, which change dynamically with the wearer's motion, making global trajectory recovery over time difficult. Considering object generalization, some rely on predefined CAD models~\cite{wen2024foundationpose, su2022zebrapose}. Though some others~\cite{ye2023diffusion,ye2024g} use differentiable rendering to generalize to different object categories, these methods are computationally expensive, sensitive to occlusions, and unstable in dynamic environments. Furthermore, current approaches underutilize egocentric priors, such as the structural coupling of the camera, body, and hands, limiting their robustness and generalization. As a result, reconstructing world-space HOI remains a major challenge, requiring methods that handle unknown objects, severe hand occlusions, and long-term spatial-temporal coherence without drift.

To address these challenges, we propose \textbf{EgoGrasp}, the first method, to the best of our knowledge, that reconstructs world-space hand–object interactions (W-HOI) from egocentric monocular videos with dynamic cameras. EgoGrasp adopts a multi-stage ``perception–generation–infilling'' framework that leverages reliable 3D cues from robust perception systems while introducing a generative motion prior to ensure temporal and global consistency.
It operates in three stages:  
(1) Preprocessing: 
We recover accurate camera trajectories, scene geometry, and hand poses from egocentric videos with off-the-shelf SOTA models~\cite{depthanything3, potamias2025wilor}. We also leverage vision foundation models to reduce the complexity of the 3D problem to 2D, enabling the robust 6DoF initialization~\cite{bargatin2025memfof, sam3dteam2025sam3d3dfyimages, li2025mvsam3d}.
(2) Body Diffusion: 
To address the instability of egocentric perspectives and frequent self-occlusions, we introduce a Body Diffusion model that synthesizes coherent body and hand motions by incorporating SMPL-X~\cite{pavlakos2019expressive} upper-body constraints.
To achieve superior accuracy, we integrate test-time optimization to enforce spatial alignment across the sequence.
(3) HOI Diffusion: 
We leverage a diffusion model to infill discrete 6DoF sequences from stage one into plausible, continuous HOI trajectories based on the reconstructed hands from stage two. By modeling natural dynamics, it provides a high-quality initialization for final test-time optimization, resulting in precise and physically-grounded 6DoF trajectories.

We validate EgoGrasp on H2O and HOI4D datasets, achieving  comparable results in world-space hand estimation and state-of-the-art results in world-space HOI reconstruction, with strong global trajectory consistency, demonstrating robustness to complex hand-object interactions and in-the-wild conditions.

Our key contributions are summarized as follows:
\begin{itemize}
    
    \item We present a comprehensive analysis of the limitations inherent in current hand pose estimation, hand–object interaction modeling, and object 6DoF tracking approaches. Building upon these insights, we introduce the task of world-space hand–object interaction (W-HOI).
    
    \item We propose a novel framework, EgoGrasp, for W-HOI reconstruction from egocentric videos. EgoGrasp features a robust pre-processing stage and two generative prior diffusion models, enabling it to produce consistent HOI trajectories in world space. Furthermore, EgoGrasp is template-free and scalable to multiple objects, making it highly flexible and generalizable.
    
    \item Extensive experiments demonstrate that EgoGrasp substantially outperforms existing methods on the H2O and HOI4D datasets, thereby establishing new state-of-the-art results for W-HOI reconstruction in real-world settings, as shown in~\cref{fig:teaser}.
\end{itemize}

\section{Related Work}

\mypara{Hand Pose Estimation.} Hand Pose Estimation has developed rapidly in recent years, with early methods primarily targeting third-person perspectives under the assumption of minimal occlusion and stable camera viewpoints. Single-hand approaches typically regress MANO~\cite{mano} model parameters~\cite{boukhayma20193d}, while two-hand methods employ implicit modeling or graph convolutions for interaction reconstruction~\cite{lee2023im2hands, li2022interacting}.
Egocentric hand estimation is crucial for teaching robots manipulation tasks from a first-person perspective, facilitating advancements in embodied intelligence and virtual reality. Existing methods~\cite{ohkawa2023assemblyhands, prakash20243d, liu2024single, wen2023hierarchical} typically reconstruct hand poses in the camera coordinate system, limiting their ability to model hand-object interactions globally. To overcome this, recent studies have explored world-space pose estimation to recover hand poses and trajectories in world coordinates. For example, Dyn-HaMR~\cite{yu2025dyn} integrates SLAM-based camera tracking with hand motion regression to achieve 4D global motion reconstruction. Similarly, HaWoR~\cite{zhang2025hawor} decouples hand motion from camera trajectories by leveraging adaptive SLAM and motion completion networks, enabling hand estimation in world coordinates.
While recent approaches have improved hand pose reconstruction, they fail to explicitly model the complex dynamics between hands and objects. Furthermore, current methods often underutilize egocentric priors, resulting in reduced robustness and generalization. To address these challenges, our EgoGrasp jointly models hand-object dynamics in world coordinates by incorporating with body-guided diffusion priors. Check \cref{tab:method-category-comparison} for differences between previous tasks. 
\mypara{Hand-Object Interaction Estimation.}
Estimating hand pose and object 6DoF is inherently challenging, especially in hand-object interaction (HOI) scenarios, where the interactions between hand and object further increase the complexity. Existing object 6DoF estimation methods can be broadly categorized as: 
(1)~template-based methods, which rely on predefined CAD models~\cite{wen2024foundationpose, su2022zebrapose} and auxiliary inputs such as segmentation masks and depth maps; 
(2)~template-free methods, which estimate the 6DoF pose without CAD models and may reconstruct the object mesh, often conditioned on RGB-D inputs and segmentation masks~\cite{lee2025any6d, wen2023bundlesdf, zhang2024omni6dpose, li2019cdpn}. Typical approaches include combining a 3D generation model with 6DoF estimation method, or per-instance methods based on differentiable rendering or implicit surface learning. However, these approaches are often computationally expensive and struggle with robustness under noise, occlusions, and dynamic conditions.

\begin{wraptable}{r}{0.45\textwidth}{ 
    \centering
    \vspace{-20pt}
    \setlength{\tabcolsep}{4pt}
\renewcommand{\arraystretch}{1.15}

\newcommand{\cmark}{\textcolor{green!55!black}{\ding{51}}} 
\newcommand{\xmark}{\textcolor{worstColor!85!black}{\ding{55}}} 
\newcommand{\mline}{\text{--}}

    \caption{
Comparison of representative tasks and world-space HOI.
\cmark: supported, \xmark: not supported, \mline: partial/ambiguous.
}
    \adjustbox{max width=1.0\linewidth}{
        \begin{tabular}{L{84pt}C{30pt}C{30pt}C{30pt}C{30pt}C{30pt}C{30pt}}
        \toprule[0.5pt]
        \makecell[l]{Category} & 
        \makecell{Ego} & 
        \makecell{Hand\\Mesh} & 
        \makecell{Obj\\6DoF} & 
        \makecell{Obj\\Mesh} & 
        \makecell{World} & 
        \makecell{Temp.} \\
        \Xhline{0.5pt}
        \makecell[l]{Exo Hand Est.} & 
        \xmark & \cmark & \xmark & \xmark & \xmark & \mline \\

        \makecell[l]{Ego Hand Est.} & 
        \cmark & \cmark & \xmark & \xmark & \xmark & \mline \\

        \makecell[l]{World Hand Est.} & 
        \mline & \cmark & \xmark & \xmark & \cmark & \cmark \\

        \makecell[l]{Camera 6DoF} & 
        \xmark & \xmark & \cmark & \mline & \xmark & \mline \\

        \makecell[l]{Camera HOI} & 
        \xmark & \cmark & \cmark & \mline & \xmark & \mline \\
        \rowcolor{bestColor}
        \makecell[l]{\textbf{W-HOI}} & 
        \cmark & \cmark & \cmark & \cmark & \cmark & \cmark \\
        \Xhline{0.5pt}
        \end{tabular}
    }
    \label{tab:method-category-comparison}
    \vspace{-10pt}
}\end{wraptable}

HOI estimation builds on 6DoF methods by adding the challenge of estimating hand pose alongside object 6DoF. Template-based methods~\cite{grady2021contactopt, liu2021semi, cao2021reconstructing} only estimate hand pose and object 6DoF, while template-free methods~\cite{ye2023diffusion, ye2024g} jointly reason about hand pose, object 6DoF, and object mesh reconstruction. Despite benefiting from joint reasoning, HOI methods face unique challenges such as severe occlusions, dynamic camera motion, and complex hand-object interactions. 
ContactOpt~\cite{grady2021contactopt} and GraspTTA~\cite{jiang2021hand} both directly optimize the contact loss by predicting or generating hand-object contact heatmaps to better construct HOI results. DiffHOI~\cite{ye2023diffusion} and G-HOP~\cite{ye2024g} also achieve object mesh reconstruction by leveraging differentiable rendering and an implicit SDF field guided by diffusion model priors. However, their computational cost is very high and their sensitivity to hand-object occlusion often results in poor temporal consistency and unstable motion reconstruction in real-world scenarios.
To resolve the aforementioned issues, we present a robust 6DoF estimation pipeline based on a 3D foundation model SAM3D~\cite{sam3dteam2025sam3d3dfyimages}, an optical flow model MEMFOF~\cite{bargatin2025memfof} and a generative diffusion model for sequence refinement. This synergy enables seamlessly consistent 6DoF tracking and long-term stable mesh reconstruction.
 \cref{tab:method-category-comparison} highlights the differences between previous tasks. 

\mypara{Motion Prior Model In Pose Estimation}
All the aforementioned hand-only estimation methods suffer from a critical limitation: the excessive number of degrees of freedom. Due to this high dimensionality, these methods are highly sensitive to various noises, causing hand orientation and positional drift, depth ambiguity, and even left–right hand misclassification. These issues fundamentally hinder stable world-space hand mesh reconstruction.
Some motion prior models have been proposed to constrain the action representation within a reasonable range.
VPoser~\cite{pavlakos2019expressive} trains a pose prior network on large-scale MoCap data to constrain SMPL-X~\cite{pavlakos2019expressive} parameters, aligning with human motion statistics. RoHM~\cite{zhang2024rohm}, LatentHOI~\cite{li2025latenthoi}, DiffHOI~\cite{ye2023diffusion}, and G-HOP~\cite{ye2024g} leverage diffusion models as priors for motion or HOI generation and reconstruction.
Similarly, we construct decoupled diffusion prior models, including a body motion diffusion model and an HOI diffusion model, to learn the upper-body pose prior and HOI prior. The upper-body pose explicitly utilizes the egocentric prior, constraining hands by body that conform to the laws of motion. 

\section{Method}
\vspace{-10pt}
\begin{figure*}[!ht]
\centering
\includegraphics[width=0.98\linewidth]{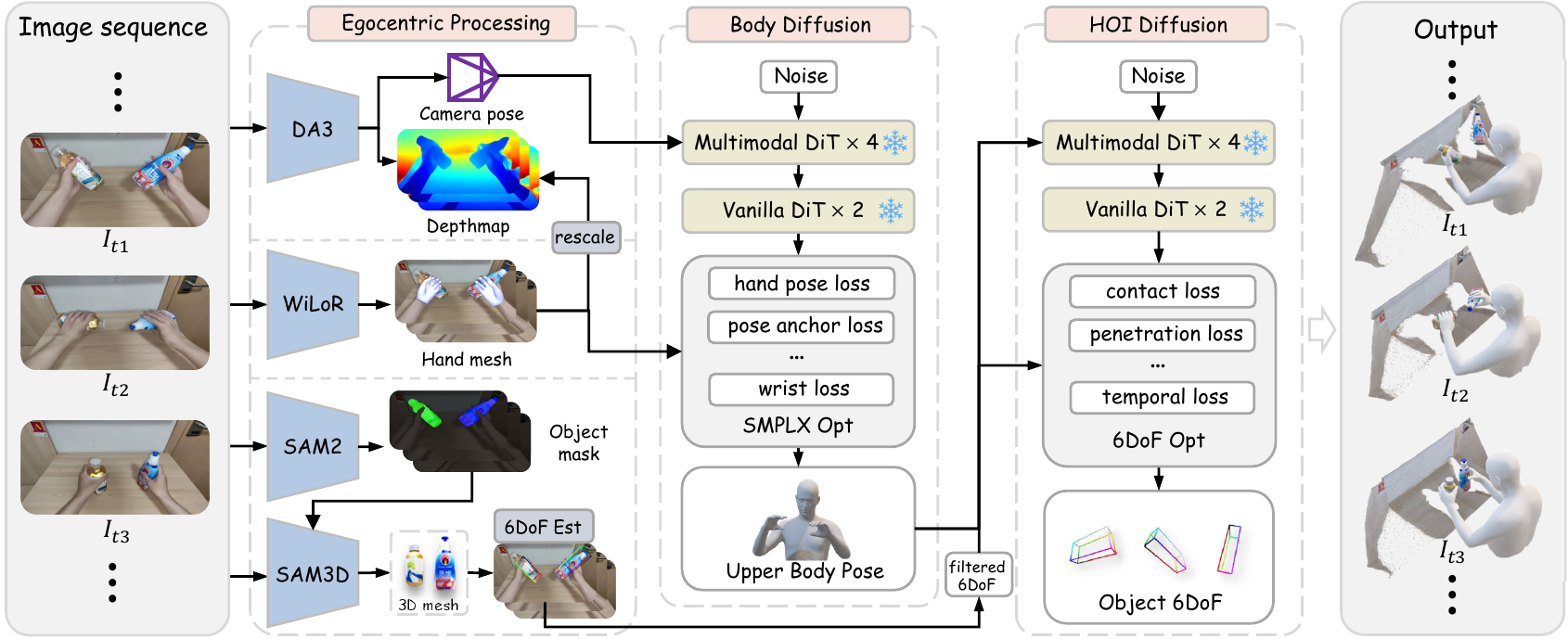}
\vspace{-5pt}
\caption{\textbf{Overview of EgoGrasp.}  We propose a three-stage method to recover world-space hand–object interaction from egocentric monocular videos with dynamic cameras: (1) extract 3D attributes with spatial perception models (for our 6DoF estimation method please see detail in \cref{fig:infiling}); (2) reconstruct upper body pose with motion prior provided by Body Diffusion; (3) Interpolate discrete 6DoF sequences with HOI Diffusion for spatial, temporal, and contact consistency.}
\label{fig:framework}
\vspace{-10pt}
\end{figure*}

\subsection{Problem Formulation}

Given an egocentric video $V \in \mathbb{R}^{T \times H \times W \times 3}$, we aim to reconstruct accurate hand-object interactions in the world-space. Different from previous methods that treat left and right hand separately, we reconstruct the upper body pose to restrict the range of dual hands: hand poses $\{\theta_l^t, \theta_r^t \in \mathbb{R}^{15 \times 3}\}_{t=0}^{T}$, body poses $\{\theta_b^t \in \mathbb{R}^{10 \times 3}\}_{t=0}^{T}$ (upper-body only), betas $\beta \in \mathbb{R}^{10}$, global orientation and translation $\{\phi_t^i,\gamma_t^i \in \mathbb{R}^3\}_{t=0}^{T}$. 
For each object, we reconstruct the canonical mesh $\mathbf{M}$ and its global trajectory $\left\{ o^t \in \mathrm{SE}(3) \right\}_{t=0}^{T}$ in world coordinates.

The proposed EgoGrasp method consists of three stages: 
1) a preprocessing stage that extracts initial 3D hand poses and objects from the video using robust off-the-self foundation models; 2) a body diffusion model that generates plausible upper body poses based on the extracted 3D attributes to constrain hand pose; and 3) an HOI diffusion model that infills the discrete 6DoF sequences from stage one into plausible and continuous sequences. An overview of the proposed method is visualized in \cref{fig:framework}.

\subsection{Egocentric Video Preprocess}
\label{subsec:3.2}
With the development of spatial intelligence, 3D tasks like point map reconstruction, parametric hand and body reconstruction, and 3D object reconstruction have also achieved significant progress, driven by advancements in foundation models and datasets.
Based on this background, we leverage off-the-shelf foundational models to process egocentric videos in three steps: global scene reconstruction~\cite{depthanything3}, hand reconstruction~\cite{potamias2025wilor}, and object reconstruction~\cite{sam3dteam2025sam3d3dfyimages}.

\mypara{Step 1. Global Scene Reconstruction.} For this step, we aim to get camera parameters and depth maps for the whole sequence. We utilize Depth-Anything3 (DA3)~\cite{depthanything3} to infer the camera intrinsics $\mathbf{K}$,  extrinsics $\mathbf{E}^t$, represented as rotation $\mathbf{R} \in \mathrm{SO}(3)$ and translation $\mathbf{T} \in \mathbb{R}^3$, along with depth maps $D^t$ for the entire video sequence. 
The global scene reconstruction of this stage serves as the global coordinate system for the subsequent transformations

\begin{wrapfigure}{r}{0.46\textwidth} 
    \centering
    \vspace{-15pt}
    \includegraphics[width=\linewidth]{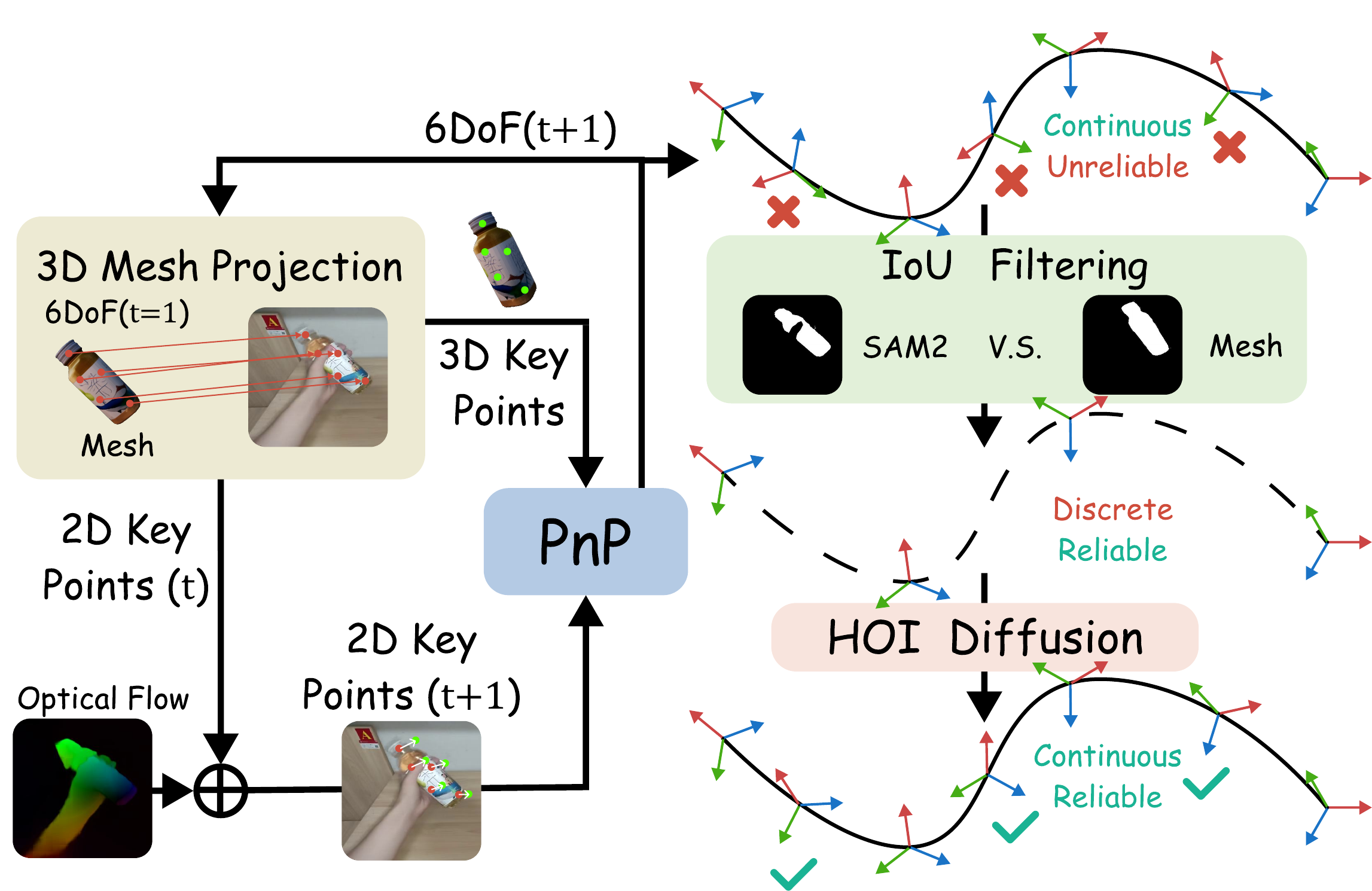}
    \vspace{-10pt}
    \caption{\textbf{Object 6DoF estimation pipeline.} (1) We use PnP and optical flow to track 6DoF based on the initial SAM3D reconstruction on the first frame. (2) We filter out unreliable 6DoFs by mask IoU. (3) HOI Diffusion from EgoGrasp infills the reliable 6DoFs.}
    \label{fig:infiling}
    \vspace{-10pt}
\end{wrapfigure}

\mypara{Step 2. Hand Reconstruction.} We reconstruct MANO~\cite{mano} parameters of both hands $\{\theta_l^t, \theta_r^t, \beta\}$ in the egoview videos by utilizing the state-of-the-art(SOTA) hand pose estimator WiLoR~\cite{potamias2025wilor}. We then use the camera focal length obtained from the first step to calculate the MANO depth: $z=f/s$. The initially estimated depth map and camera translation from step 1 are aligned to the metric scale by multiplying a global scale factor. This factor is computed as the mean ratio between the depth values of the MANO model and the corresponding regions in the estimated depth map.

\mypara{Step 3. Object Reconstruction.}
We propose a robust pipeline to reconstruct 6DoF poses of open-vocabulary objects, consisting of mesh initialization, temporal tracking, and trajectory refinement.
First, to obtain a high-fidelity template mesh $\mathbf{M}$, we employ SAM3D~\cite{sam3dteam2025sam3d3dfyimages} using its multi-view implementation~\cite{li2025mvsam3d} to enrich geometric details. This process yields the canonical shape $\mathbf{M}$ and the initial pose. 
To propagate this pose across the video, we utilize MEMFOF~\cite{bargatin2025memfof}, an optical-flow-based tracker. By projecting 3D keypoints from $\mathbf{M}$ and tracking their 2D displacements, we solve the Perspective-n-Point (PnP) problem to recover a preliminary 6DoF trajectory.
The initial tracking may suffer from drift, especially during hand-object occlusions. Instead of relying solely on frame-wise tracking, we introduce a refinement stage. We first filter unreliable frames by checking the IoU between the SAM2 mask and the projected mesh mask, marking low-IoU frames as invalid (zero-padded). This sparse sequence $\mathbf{o}_{filtered}$ is then processed by our HOI Diffusion model. Leveraging generative priors, the model effectively infills the missing segments and rectifies the trajectory, producing physically plausible and continuous 6DoF motion (see \cref{fig:infiling}). Further details of our HOI diffusion are in \cref{sec:hoi_diff}.

\subsection{Body Diffusion Model}
After obtaining the initial results from the preprocessing stage, we need a prior model with HOI knowledge to impose physical constraints and perform temporal completion. However, HOI datasets that include full-body poses are extremely limited. Training a unified model to simultaneously handle hand pose estimation and object 6DoF predictions under such constraints introduces substantial pose bias, which impedes the ability of the model to learn meaningful body motion priors.
To address this, we propose to decouple the upper body pose estimation from the object 6DoF, training two separate networks, a Body Diffusion Model $\mathcal{W}$ and an HOI Diffusion Model $\mathcal{H}$, as shown in \cref{fig:framework}. 
We first utilize $\mathcal{W}$ to generate the upper-body pose, leveraging the body's limited reach range to constrain the hand's movement region, rather than treating both hands as independently translatable objects. 

We additionally condition the model with features $\mathbf{c}^t$, extracted from a transformer condition encoder $\phi$ only from the central pupil frame (CPF) inter-frame transformations $\Delta\mathbf{T}_{\mathrm{cpf}}^{t-1\rightarrow t}$. 
Following Egoallo~\cite{yi2025estimating}, the CPF serves as a canonical reference coordinate system to bridge the SMPL-X local pose and the world space. Specifically, the origin of the CPF is anchored at the interocular midpoint of the SMPL-X model, while its axes are strictly aligned with the head orientation.
The formulation of the body diffusion model's conditions is given as follows: 
{
\begin{align}
\Delta\mathbf{T}_{\mathrm{cpf}}^{t\rightarrow t+1}
= \left(\mathbf{E}^{t}\right)^{-1}\mathbf{E}^{t+1} \in \mathrm{SE}(3), 
\mathbf{c}^t = \phi\Big(
\Delta\mathbf{T}_{\mathrm{cpf}}^{t-1\rightarrow t}
\Big).
\end{align}
}

Let $\mathbf{z}_0^{\,1:T} = \theta_{\mathrm{body}}^{\,1:T}$ denote the ground truth SMPL-X body parameters. We formulate the body motion estimation as a conditional Denoising Diffusion Probabilistic Model (DDPM)~\cite{ho2020denoising}. Our body diffusion model $\mathcal{W}$ directly reconstructs the clean parameters $\mathbf{z}_0^{\,1:T}$ from the noisy state $\mathbf{z}_{t_d}^{\,1:T}$ at timestep $t_d$, conditioned on the features $\mathbf{c}^{\,1:T}$. The model is trained by minimizing the reconstruction error:
\begin{align}
    \mathcal{L}_{\mathrm{W}} = \mathbb{E}_{t_d,\;\mathbf{z}_0,\;\boldsymbol{\epsilon}} \left[ \left\| \mathcal{W}\big(\mathbf{z}_{t_d}^{\,1:T},\;\mathbf{c}^{\,1:T},\;t_d\big) - \mathbf{z}_0^{\,1:T} \right\|_2^2 \right].
\end{align}

During inference, the final predicted body parameters $\widehat{\theta}_{\mathrm{body}}^{\,1:T}$ are obtained by denoising a sequence iteratively sampled from pure noise $\mathcal{N}(\mathbf{0},\mathbf{I})$.

\mypara{Test-time Optimization.}
  At test time, we refine SMPL-X parameters with several objectives to ensure realistic and physically plausible body motion. 
  (1)~Pose anchor loss $\mathcal{L}_{body}$ prevents excessive drift by keeping the optimized body configuration close to the predicted one, especially for arm and hand pose parameters.
  (2)~Hand pose loss $\mathcal{L}_{pose}$ aligns the optimized left and right hand poses with the hand-pose predictions from WiLoR.
  (3)~Hand 2D keypoint loss $\mathcal{L}_{kp2d}$ enforces image-space consistency by matching the projected 2D coordinates of our 3D hand joints to the
  corresponding WiLoR keypoints on jointly visible joints.
  (4)~Wrist loss $\mathcal{L}_{wrist}$ enforces consistency of wrist 6DoF with WiLoR by constraining both wrist position and wrist orientation in the
  CPF coordinates.
  Please refer to the Supp.Mat for additional details.

\subsection{HOI Diffusion Model}
\label{sec:hoi_diff}Our HOI diffusion model 
$\mathcal{H}$ is designed to generate plausible object 6DoF trajectories by leveraging reliable object predictions and human motion priors. It takes as input the initial reliable object 6DoF predictions $\mathbf{o}^t_{filtered}$, which are filtered by mask IoU and represented in wrist coordinates, alongside the conditions features $\mathbf{m}^t$  produced by a transformer condition encoder $\phi_{obj}$. Specifically, the conditioning variables comprise: 
(1) the reliable object 6DoFs $\mathbf{o}^t_{filtered}$ in wrist coordinates; 
(2) a coarse translation $\mathbf{t}^t_{coarse}$ derived by projecting the masked depth using camera intrinsics and subsequently transformed into wrist coordinates;
(3) a one-hot grasping label $\mathbf{l}^t_{grasp}$ determined by hand-to-object distances; 
(4) the hand poses $\theta^t_{obj}$ and hand joint locations $\mathbf{x}^t_{obj}$ in wrist coordinates generated by the body diffusion model $\mathcal{W}$.
We train $\mathcal{H}$ using a masked reconstruction objective, to recover the randomly masked ground truth 6DoF trajectories. This enables effective training on existing HOI datasets while preserving the body motion priors learned from large-scale full-body datasets. The inference formula of HOI diffusion is as follows:

\begin{gather}
    {\mathbf{m}}_{j}^{1:T} = \phi_{obj}\big({\mathbf{o}}_{filtered}^{1:T}, \mathbf{t}^t_{coarse}, \mathbf{l}_{grasp}^{1:T}, \mathbf{x}_{obj}^{1:T}, \theta_{obj}^{1:T} \big),\\
    \widehat{\mathbf{o}}_{t_d-1}^{1:T} = \mathcal{H}\big(\mathbf{o}_{t_d}^{1:T}, {\mathbf{m}}_{}^{1:T}, \mathbf{o}_{filtered}^{1:T}, t_d\big),t_d \in [1, 1000],
\end{gather}

where $\mathbf{o}_{}$ denotes the 6DoF of each object. We train the HOI diffusion model $\mathcal{H}$ similar to the body diffusion model $\mathcal{W}$. 

By looping through each object outside the HOI diffusion $\mathcal{H}$, multi-object interactions can be achieved, please check Supp.Mat. for details. Note that the object meshes $\mathbf{M}$ are obtained as described in \ref{subsec:3.2}.

\mypara{Test-time Optimization.}
  At test time, we perform a lightweight, fully differentiable optimization to refine object poses.
  The objective jointly enforces spatial accuracy and temporal smoothness for realistic and physically plausible object motion.
  (1)~Object anchor loss $\mathcal{L}_{obj}$ prevents excessive drift by regularizing the refined object 6DoF toward the HOI diffusion prediction.
  (2)~Contact loss $\mathcal{L}_{contact}$ enforces near-zero relative motion at contact by penalizing temporal displacement of grasping hand contact
  points in the object local frame, effectively reducing unnatural sliding.
  (3)~Penetration loss $\mathcal{L}_{penetration}$ prevents non-physical interpenetration by penalizing hand points that violate a safety margin to the
  object surface during complex HOI sequences.
  (4)~Temporal loss $\mathcal{L}_{temporal}$ regularizes object motion over time by penalizing unstable velocity and acceleration in both rotation and translation, improving trajectory smoothness.
  For additional details please refer to the Supp.Mat.

\section{Experiments}

\subsection{Implementation Details \& Metrics}

We evaluate hand pose estimation and object 6DoF estimation on two egocentric HOI datasets, the H2O~\cite{Kwon_2021_ICCV} and HOI4D~\cite{Liu_2022_CVPR}. Following Dyn-HaMR~\cite{yu2025dyn} and HaWoR~\cite{zhang2025hawor}, the metrics employed for hand estimation evaluation included  World Mean Per Joint Position Error (W-MPJPE, $mm$), World-aligned Mean Per Joint Position Position Error (WA-MPJPE, $mm$), and Procrustes-aligned Mean Per Joint Position Error (PA-MPJPE, $mm$). For the 6DoF evaluation, we employ Relative Rotation Error (RRE, $^\circ$) and Relative Translation Error (RTE, $mm$) across world, local, and wrist coordinate systems. Additionally, Chamfer Distance (CD, $cm^2$) in the wrist coordinates and Penetration Depth (PD, $mm$) are utilized to assess the geometric fidelity and physical plausibility of the HOI results. World-space metrics are computed over segments of 128 frames, where W-MPJPE involved aligning only the first two frames, whereas WA-MPJPE aligned the entire segment, both using Procrustes Alignment.

We train the model using PyTorch with 2 NVIDIA H20 GPUs at a learning rate of 2.5e-4, employing AdamW optimizer and cosine annealing. The body diffusion was trained on AMASS~\cite{AMASS:ICCV:2019}, 100STYLE~\cite{mason2022real}, GRAB~\cite{GRAB:2020}, PA-HOI~\cite{wang2025pa}, and HIMO~\cite{lv2024himo} datasets; HOI diffusion was trained on GRAB~\cite{GRAB:2020}, PA-HOI~\cite{wang2025pa}, and  HIMO~\cite{lv2024himo} datasets. Training sequences were sampled at 30 FPS with random lengths ranging from 64 to 256 frames. During test-time optimization, we used learning rates of 2.5e-4, 2.5e-4, and 1.0e-4 to optimize hand pose, body pose, and beta parameters, respectively, and performed a total of 50 optimization steps using AdamW and cosine annealing. For inference, we applied DDIM~\cite{song2020denoising} sampling with 200 steps, and downsampled sequences by 3 at preprocessing stage.

\subsection{World-Space Hand Pose Estimation}
We demonstrate the superior reconstruction quality of EgoGrasp on the world-space hand pose by comparing it with the state-of-the-art Dyn-HaMR~\cite{yu2025dyn} and HaWoR~\cite{zhang2025hawor} methods.

\mypara{Quantitative Comparisons.}
\cref{tab:exp_hand} presents the quantitative results of EgoGrasp and other competing methods on the H2O and HOI4D datasets. As a complicated W-HOI framework, EgoGrasp achieves competitive results, reaching optimal or near-optimal performance compared to newly proposed SOTA hand-specific methods Dyn-HaMR and HaWoR. 
HaWoR performs the worst, with its metrics falling significantly behind both EgoGrasp and Dyn-HaMR. Furthermore, Dyn-HaMR face a performance drop (in PA-MPJPE) when  processing long-range HOI trajectories within the HOI4D dataset.

{
\begin{table}[h]
\centering
\caption{Hand pose evaluation on H2O and HOI4D datasets. For each metric, we use background colors to indicate the \besttxt{best}, \secondtxt{second best}.}
\setlength{\tabcolsep}{1.5mm} 

\resizebox{\linewidth}{!}{
\begin{tabular}{l | c c c | c c c}
\toprule
\multirow{2}{*}{Method} & \multicolumn{3}{c|}{H2O} & \multicolumn{3}{c}{HOI4D} \\
\cmidrule(lr){2-4} \cmidrule(l){5-7}
& W-MPJPE $\downarrow$ & WA-MPJPE $\downarrow$ & PA-MPJPE $\downarrow$ & W-MPJPE $\downarrow$ & WA-MPJPE $\downarrow$ & PA-MPJPE $\downarrow$ \\ 
\midrule
Dyn-HaMR (CVPR25) & \best{5.75} & \second{46.37} & \best{16.74} & \second{9.83} & \best{190.68} & \second{63.12} \\
HaWoR (CVPR25)  & \second{5.77} & \worst{113.39} & \worst{30.75} & \worst{9.04} & \worst{196.09} & \worst{69.20} \\
EgoGrasp      & \worst{6.84} & \best{40.93} & \second{18.92} & \best{8.61} & \second{192.06} & \best{47.29} \\
\bottomrule
\end{tabular}
}
\label{tab:exp_hand}
\end{table}
}

\begin{figure*}[!h]
    \centering
    \includegraphics[height=16cm, width=\textwidth, keepaspectratio]{img/exp_hand.pdf}
    \caption{World-space hand pose visualizations on the H2O dataset (top two sequences) and the HOI4D dataset (bottom two sequences).}
    \label{fig:hand_pose}
\end{figure*}

\mypara{Qualitative Comparisons.}
\cref{fig:hand_pose} shows a qualitative comparison of HaWoR, Dyn-HaMR, and EgoGrasp, demonstrating that EgoGrasp achieves superior performance in both fine-grained hand manipulation on the H2O dataset and long-term motion trajectories on the HOI4D dataset. 
As illustrated in the first two sequences, EgoGrasp accurately reconstructs subtle hand movements, whereas the competing methods exhibit significant drift. Notably, HaWoR suffers from severe trajectory deviations, failing to produce a physically plausible path. In the subsequent sequences from HOI4D, HaWoR continues to show pronounced drifting and pose artifacts, while Dyn-HaMR even fails at hand-side identification (left vs. right). Furthermore, both baselines yield inaccurate world-space trajectories. In contrast, by incorporating a body pose constraint, EgoGrasp ensures kinematic consistency with natural human motion, resulting in substantially more precise and stable reconstructions.

\subsection{World-Space Hand-Object Interaction Estimation}
To further evaluate the effectiveness of EgoGrasp in the W-HOI task, we perform a comparative analysis against 3 baselines: 6DoF method Any6D \cite{lee2025any6d} + WiLoR and GenPose2 \cite{zhang2024omni6dpose} + WiLoR, as well as a feed-forward HOI model HORT \cite{chen2025hort}.
We omit comparisons with per-instance optimization methods (e.g., differentiable rendering, implicit surface learning) due to their high time overhead and lack of scalability for large-scale HOI data.
The 3 baselines differ in their output representations: HORT is a feed-forward model that only yields object point clouds based on WiLoR output, while GenPose2 is restricted to bounding box estimation and cannot generate meshes. 
Since HORT does not directly provide 6DoF poses, we employ the Iterative Closest Point (ICP) algorithm to register its predicted point clouds with the ground-truth meshes, thereby deriving the temporal 6DoF poses required for quantitative evaluation. It is important to note that the metrics in wrist coordinates are calculated by converting 6DoF or mesh to the corresponding hand wrist coordinates. For GenPose2, Any6D, and HORT, the hand results of WiLoR are used for calculating metrics in wrist coordinates.

{
\begin{table}[h]
\centering
\caption{Object 6DoF estimation and HOI evaluation on H2O and HOI4D datasets. For each metric, we use background colors to indicate the \besttxt{best} and \secondtxt{second best}.}
\label{tab:exp_obj}
\setlength{\tabcolsep}{5.5mm} 

\resizebox{\linewidth}{!}{
\begin{tabular}{l cccccccc}
\toprule
\multirow{2}{*}{\textbf{Method}} & \multicolumn{2}{c}{\textbf{Local}} & \multicolumn{2}{c}{\textbf{Global}} & \multicolumn{3}{c}{\textbf{Wrist}} \\
\cmidrule(lr){2-3} \cmidrule(lr){4-5} \cmidrule(lr){6-8}
& RRE $\downarrow$ & RTE $\downarrow$ & RRE $\downarrow$ & RTE $\downarrow$ & RRE $\downarrow$ & RTE $\downarrow$ & CD $\downarrow$ & PD $\downarrow$ \\
\midrule

\rowcolor[HTML]{F3F3F3} \multicolumn{9}{l}{\textit{Results on H2O Dataset}} \\ 
GenPose2 (ECCV24) & \second{28.88} & \best{72.93} & \second{28.53} & \best{75.75} & \second{33.19}	& 74.19 & - & -\\
Any6D (CVPR25)  & 33.40 & 85.43 & 33.77 & 89.62 & 37.27 &	\worst{80.82}  & \second{52.51} & \second{3.18}\\
HORT (ICCV25) & \worst{35.09} & \worst{116.42} & \worst{35.31} & \worst{125.82} & \worst{37.75} & \best{56.22} & \best{22.98} & 10.46\\
EgoGrasp & \best{18.62} &	\second{77.68} &	\best{18.71} &	\second{77.84} &	\best{23.08} &	\second{58.30} & 52.56 & \best{0.74}\\

\midrule[0.6pt]
\rowcolor[HTML]{F3F3F3} \multicolumn{9}{l}{\textit{Results on HOI4D Dataset}} \\
GenPose2 (ECCV24)  & \second{21.15} & 188.75 & \second{23.16} & \second{178.94} & \second{47.15} & 124.84 & - & -\\
Any6D (CVPR25) & \worst{36.92} & \worst{201.28} & \worst{37.60} & 187.01 & 55.16 & \worst{187.03} & 499.70 & \best{0.46}\\
HORT (ICCV25) & 29.04 & \second{166.36} & 34.67 & \worst{256.26} & \worst{54.63} & \second{118.90} & \best{114.03} & 5.56\\
EgoGrasp & \best{14.33} & \best{124.69} & \best{14.17} & \best{135.80} & \best{34.87} & \best{116.66} & \second{235.53} & \second{0.64}\\

\bottomrule
\end{tabular}
}
\end{table}
}

\begin{figure*}[!ht]
    \centering
    \includegraphics[height=12cm, width=\textwidth, keepaspectratio]{img/exp_hoi.pdf}
    \caption{World-space hand-object interaction visualizations on  H2O (top two sequences) and HOI4D dataset (bottom two sequences).}
    \label{fig:obj_pose}
\end{figure*}

\mypara{Quantitative Comparisons.}
\cref{tab:exp_obj} presents the performance of various methods on the H2O and HOI4D datasets for object 6DoF tracking. Here, all other methods except EgoGrasp use ground-truth camera extrinsics to transform results from the camera coordinate system to the world coordinate system, whereas EgoGrasp uses camera extrinsics predicted by DA3 and poses predicted by body diffusion to transform SMPL-X into world coordinates.  

The experimental results demonstrate that EgoGrasp achieves substantial improvements in the RRE and RTE metrics across local, global, and wrist coordinate systems. In particular, the significant lead in the RRE metric underscores the robustness of EgoGrasp in modeling complex hand-object interactions, whereas other comparative methods remain limited to simple operations such as object translation. Although GenPose2 exhibits a marginal advantage in the RTE for local and global coordinates on the H2O dataset, the performance of this method suffers from severe degradation in the rotation and wrist coordinate systems. While HORT attains the leading performance in Chamfer Distance (CD), it functions as a single-frame method that only outputs point clouds, thus lacking temporal consistency and a complete mesh structure. This limitation leads to the highest Penetration Depth (PD) among all evaluated models, which compromises physical plausibility. In contrast, EgoGrasp maintains superior values in both the CD and PD metrics.

\mypara{Qualitative Comparisons.}
As shown in \cref{fig:obj_pose}, only EgoGrasp successfully reconstructs the object trajectory while simultaneously achieving accurate hand pose estimation. This superiority benefits from our body and HOI diffusion model, which jointly optimizes the initial hand pose and 6DoF estimations.

The initial two sequences demonstrate that GenPose2 is restricted to bounding box estimation, whereas the mesh generated by Any6D suffers from near-total failure, accompanied by substantial errors in the estimated 6DoF pose. Furthermore, the point cloud output produced by HORT fails to preserve the structural integrity of the object, manifesting significant divergence. In contrast, EgoGrasp achieves 6DoF estimation that remains fundamentally consistent with the ground truth while maintaining high-quality mesh reconstruction.

Regarding the subsequent two sequences, the three baseline methods exhibit varying degrees of trajectory jitter and physically implausible hand-object contact. EgoGrasp stands out by achieving robust reconstruction of long-range hand-object interactions within world-space coordinates.

\subsection{Ablation Studies}
To demonstrate the contribution of EgoGrasp component, we implemented several variants. 

\mypara{Ablations on Hand.} 
Specifically, the variants ``\textit{w/o} $\mathcal{L}_{kp2d}$, \textit{w/o} $\mathcal{L}_{pose}$, \textit{w/o} $\mathcal{L}_{wrist}$'' denote the removal of the corresponding loss terms during the SMPL-X test-time optimization process.
\cref{tab:exp_ablation_hand} reports hand-pose ablation results on the H2O and HOI4D datasets. We observe that all variants except EgoGrasp exhibit significant declines in certain metrics, while EgoGrasp consistently achieves the best or competitive results. Notably, $\mathcal{L}_{wrist}$ demonstrates its important role in hand positioning, while $\mathcal{L}_{kp2d}$ and $\mathcal{L}_{pose}$ further optimize the poses based on it. The marginal decline of EgoGrasp in specific metrics (e.g., WA-MPJPE) compared to certain ablation variants is attributed to the synergistic interaction and trade-off between different loss terms, which prevents the optimization from over-fitting to a single metric. Overall, EgoGrasp demonstrates the most robust comprehensive performance.

{
\begin{table}[t]
\centering
\caption{Ablations study of hand pose estimation on H2O and HOI4D datasets. For each metric, we use background colors to indicate the \besttxt{best}, \secondtxt{second}, and \worsttxt{worst}.}
\setlength{\tabcolsep}{3.5mm} 

\resizebox{\linewidth}{!}{
\begin{tabular}{l | c c c | c c c}
\toprule
\multirow{2}{*}{Method} & \multicolumn{3}{c|}{H2O} & \multicolumn{3}{c}{HOI4D} \\
\cmidrule(lr){2-4} \cmidrule(l){5-7}
& W-MPJPE $\downarrow$ & WA-MPJPE $\downarrow$ & PA-MPJPE $\downarrow$ & W-MPJPE $\downarrow$ & WA-MPJPE $\downarrow$ & PA-MPJPE $\downarrow$ \\ 
\midrule

\textit{w/o} $\mathcal{L}_{kp2d}$     & \worst{9.60} & 47.25 & 22.06 & 8.97 & \best{189.41} & 48.32 \\

\textit{w/o} $\mathcal{L}_{pose}$     & 8.76 & \second{42.84} & \second{20.32} & \worst{10.44} & 195.31 & \second{48.30} \\

\textit{w/o} $\mathcal{L}_{wrist}$     & \second{8.11} & \worst{238.20} & \worst{50.66} & \second{8.78} & \worst{321.84} & \worst{92.69} \\

EgoGrasp      & \best{6.84} & \best{40.93} & \best{18.92} & \best{8.61} & \second{192.06} & \best{47.29} \\

\bottomrule
\end{tabular}
}
\label{tab:exp_ablation_hand}
\end{table}
}

{
\begin{table}[t]
\centering
\caption{Ablation study of object 6DoF estimation on H2O and HOI4D datasets. 
For each metric, we use background colors to indicate the \besttxt{best}, \secondtxt{second}, and \worsttxt{worst}.}
\label{tab:exp_ablation_obj}
\setlength{\tabcolsep}{5.5mm} 

\resizebox{\linewidth}{!}{
\begin{tabular}{l cccccccc}
\toprule
\multirow{2}{*}{\textbf{Method}} & \multicolumn{2}{c}{\textbf{Local}} & \multicolumn{2}{c}{\textbf{Global}} & \multicolumn{3}{c}{\textbf{Wrist}} \\
\cmidrule(lr){2-3} \cmidrule(lr){4-5} \cmidrule(lr){6-8}
& RRE $\downarrow$ & RTE $\downarrow$ & RRE $\downarrow$ & RTE $\downarrow$ & RRE $\downarrow$ & RTE $\downarrow$ & CD $\downarrow$& PD $\downarrow$ \\
\midrule

\rowcolor[HTML]{F3F3F3} \multicolumn{9}{l}{\textit{Results on H2O Dataset}} \\

\textit{w/o} $\mathcal{L}_{contact}$ & \best{18.41} & \worst{105.67} & \best{18.42} & \worst{107.10} & \best{22.91} & \worst{77.27} & \worst{54.89} & \worst{0.98} \\

\textit{w/o} $\mathcal{L}_{temporal}$ & 19.12 & \second{78.81} & 19.19 & \second{79.27} & 23.45 & \best{57.11} & \best{47.61} & 0.93 \\

\textit{w/o} $\mathcal{L}_{penetration}$  & 18.88 & 78.85 & 18.95 & 79.28 & 23.28 & \second{57.20} & \best{47.61} & 0.93\\

\textit{w/o} Opt.  & \worst{19.48} & 102.57 & \worst{19.49} & 104.16 & \worst{24.20} & 68.98 & \second{49.16} & \second{0.81}\\

EgoGrasp & \second{18.62} &	\best{77.68} & \second{18.71} &	\best{77.84} &	\second{23.08} &	58.30 & 52.56 & \best{0.74}\\

\midrule[0.6pt]
\rowcolor[HTML]{F3F3F3} \multicolumn{9}{l}{\textit{Results on HOI4D Dataset}} \\

\textit{w/o} $\mathcal{L}_{contact}$ & \second{15.22} & \worst{144.85} & 15.22 & \worst{155.15} & 34.99 & \worst{132.88} & \worst{251.34} & 0.87\\

\textit{w/o} $\mathcal{L}_{temporal}$ & \worst{15.37} & 126.75 & \worst{15.25} & 137.32 & \best{34.81} & \second{117.26} & \second{229.23} & \worst{0.88}\\

\textit{w/o} $\mathcal{L}_{penetration}$  & 15.27 & \second{126.71} & \second{15.15} & \second{137.23} & \second{34.85} & 117.28 & \best{228.77} & \worst{0.88}\\

\textit{w/o} Opt.  & 15.25 & 142.32 & 15.18 & 153.36 & \worst{35.00} & 128.39 & 245.52 & \second{0.78}\\

EgoGrasp & \best{14.33} & \best{124.69} & \best{14.17} & \best{135.80} & 34.87 & \best{116.66} & 235.53 & \best{0.64}\\

\bottomrule
\end{tabular}
}
\end{table}
}

\mypara{Ablations on Object.}
 ``\textit{w/o} $\mathcal{L}_{contact}$, \textit{w/o} $\mathcal{L}_{temporal}$, \textit{w/o} $\mathcal{L}_{penetration}$'' refer to the exclusion of respective loss terms during object 6DoF test-time optimization. ``\textit{w/o} Opt.'' represents the raw results obtained directly from HOI diffusion without any post-optimization.
\cref{tab:exp_ablation_obj} presents the object-6DoF ablation results on H2O and HOI4D. The performance trends mirror those observed in the hand-pose ablations: most variants show a substantial drop in specific indicators. For instance, removing $\mathcal{L}_{contact}$ leads to a severe decline in RTE and CD, highlighting its necessity for accurate translation. The absence of $\mathcal{L}_{temporal}$ results in a significant increase in RRE, underscoring its importance for rotational consistency. Furthermore, $\mathcal{L}_{penetration}$ effectively minimizes physical violations, as evidenced by the degraded PD metric when it is removed. 
While the ``\textit{w/o} Opt.'' yields acceptable results, which validates the strong infilling capacity of our HOI diffusion, it still underperforms across all metrics compared to the full EgoGrasp, proving the necessity of test-time optimization. 
Although EgoGrasp slightly trails in isolated metrics due to the balancing effect of multiple constraints, it remains best in overall metrics.

\section{Conclusion}

We introduced EgoGrasp, the first method to reconstruct world-space hand–object interactions (W-HOI) from egocentric monocular videos captured by dynamic cameras. Our multi-stage framework integrates a robust pre-processing pipeline built on vision foundation models, an body-guided prior diffusion model for hand estimation, and a template-free object 6DoF estimation pipline based on an HOI diffusion model. EgoGrasp yields accurate, physically plausible, and temporally coherent W-HOI trajectories that generalize beyond single-object and template constraints. Experiments on H2O and HOI4D datasets demonstrate its advanced performance under long-range motion and severe hand-object occlusion. 
Looking ahead, we plan to explore more streamlined feed-forward architectures for modeling world-space human–object interaction, with the goal of reducing reliance on  preprocessing while preserving accuracy and robustness.

\clearpage

%
%
\bibliographystyle{splncs04}
\bibliography{main}

\appendix
\section{Transformations}

\begin{wrapfigure}{r}{0.46\textwidth} 
    \centering
    \vspace{-15pt}
    \includegraphics[width=\linewidth]{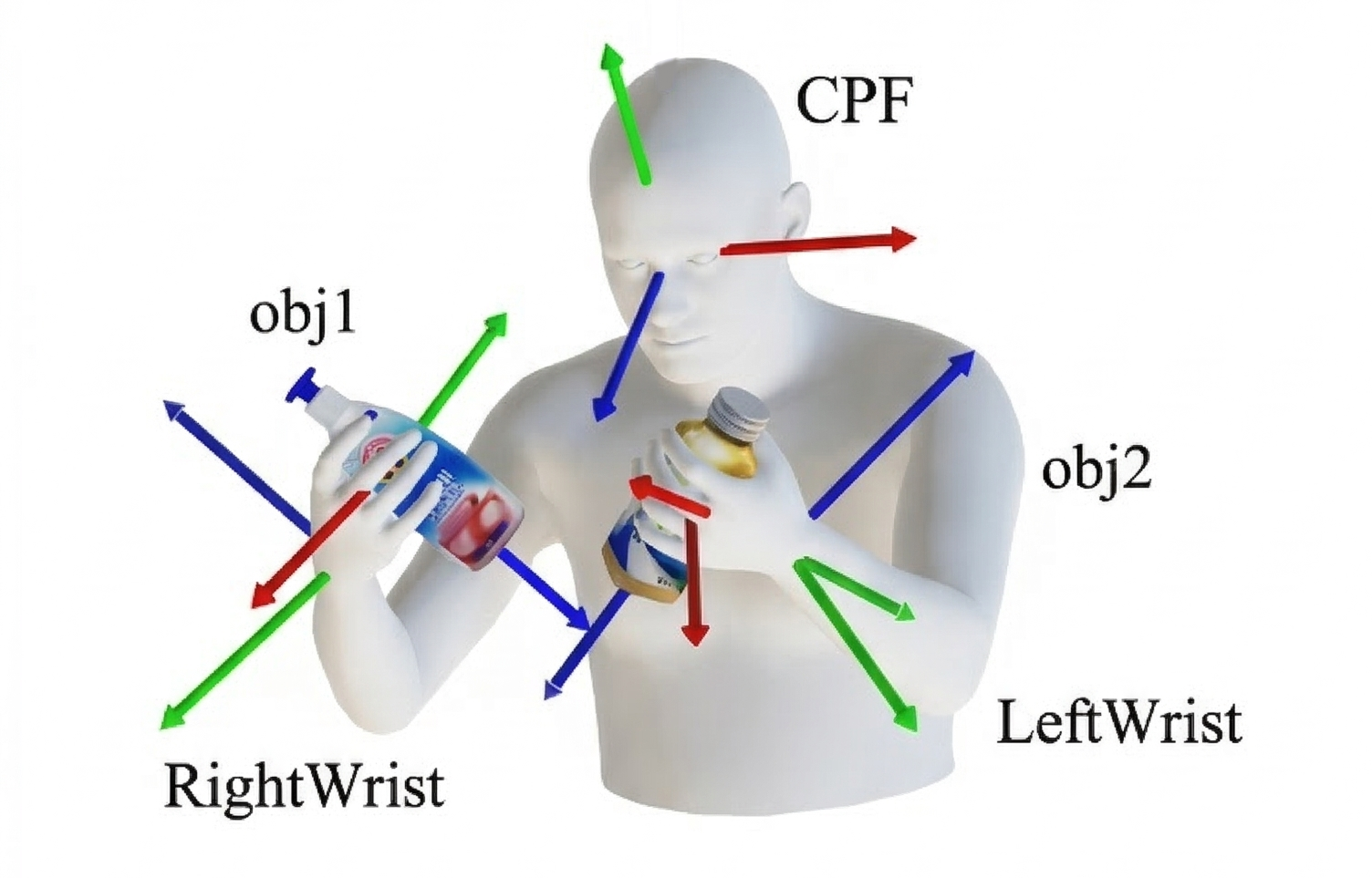}
    \vspace{-10pt}
    \caption{\textbf{Transformations visualization.} The coordinate axes in the figure represent the corresponding coordinate system transformations.}
    \label{fig:CPF}
    \vspace{-10pt}
\end{wrapfigure}

\cref{fig:CPF} illustrates the upper-body SMPL-X mesh generated by EgoGrasp, together with the visualizations of three coordinate frames: LeftWrist, RightWrist, and CPF. The rigid transformations among these coordinate frames are used as conditions and inputs for the body diffusion model and SMPL-X test-time optimization, guiding EgoGrasp to synthesize upper-body poses that are consistent with the egocentric viewpoint priors. 
It also shows the object coordinates, note that HOI Diffusion infills object 6DoF in wrist coordinates, then it can be transformed into CPF coordinates.

\section{6DoF Tracking \& HOI Diffusion Infilling}

This note describes Step 3 (\emph{Object Reconstruction}) of the main paper. Let the video contain frames $t=1,\dots,T$, with RGB image $I_t$, depth map $D_t$, camera intrinsics $K_t\in\mathbb{R}^{3\times 3}$, and SAM2 object mask $S_t\subset\Omega$, where $\Omega$ denotes the image domain. 
And the object pose in camera coordinates is
\begin{equation}
T_t^{o\rightarrow c}=
\begin{bmatrix}
R_t & \mathbf{t}_t\\
\mathbf{0}^{\top} & 1
\end{bmatrix}
\in SE(3),
\end{equation}
where $R_t\in SO(3)$ and $\mathbf{t}_t\in\mathbb{R}^{3}$.

\mypara{Anchor-frame reconstruction.}
An anchor frame $a$ is selected and provides an image $I_t$, a mask $S_t$, and a masked pointmap $\mathbf{P}_t$.
These observations are fed to SAM3D, which outputs a canonical mesh $\mathbf{M}$, an anchor-frame pose, and a mesh scale $\mathbf{s}$.
And we use the scaled mesh $\mathbf{M}_{\mathbf{s}}$.

\mypara{Pose propagation by MEMFOF and PnP.}
Starting from the anchor pose, 6DoF tracking is performed on sampled frames and then interpolated to the full video. Let
$F_{t\rightarrow t+1}:\Omega\rightarrow\mathbb{R}^2$
denote the optical flow from frame $t$ to frame $t+1$ predicted by MEMFOF. Given the current pose $T_t^{o\rightarrow c}$, surface points $\{\mathbf{X}_i\}$ are sampled on $\mathbf{M}_{\mathbf{s}}$, projected to frame $t$, and kept only if they fall inside $S_t$. Then the optical flow transports each visible point to the next frame.
\begin{equation}
\mathbf{x}_i^{t}=R_t\mathbf{X}_i+\mathbf{t}_t,\qquad
\mathbf{u}_i^{t}=\pi_{K_t}(\mathbf{x}_i^{t}), \qquad
\tilde{\mathbf{u}}_i^{t+1}
=
\mathbf{u}_i^{t}+F_{t\rightarrow t+1}(\mathbf{u}_i^{t}).
\end{equation}
where $\pi_{K_t}(\cdot)$ denotes standard perspective projection under intrinsics $K_t$. This gives candidate 2D--3D correspondences:
\begin{equation}
\mathcal{C}_{t\rightarrow t+1}
=
\left\{
\bigl(\mathbf{X}_i,\tilde{\mathbf{u}}_i^{t+1}\bigr)
\right\}.
\end{equation}
The next pose is then estimated by RANSAC-PnP followed by refinement:
\begin{equation}
T_{t+1}^{o\rightarrow c}
=
\arg\min_{T\in SE(3)}
\sum_{i\in\mathcal{I}_{t+1}}
\left\|
\pi_{K_{t+1}}\!\left(T\mathbf{X}_i\right)
-
\tilde{\mathbf{u}}_i^{t+1}
\right\|_2^2,
\end{equation}
where $\mathcal{I}_{t+1}$ indexes the 2D--3D correspondences. Propagating forward and backward from the anchor yields all the frames.

\mypara{Mask-IoU reliability.}
For each frame, the current mesh is rasterized to a binary mask. And hand-occluded pixels are removed by hand mask $H_t$.

\begin{equation}
\widehat{S}_t=\mathcal{R}(\mathbf{M}_{\mathbf{s}},T_t^{o\rightarrow c},K_t), \quad \quad
\widehat{S}_t^{\mathrm{vis}}=\widehat{S}_t\setminus H_t,
\end{equation}
where $\mathcal{R}(\cdot)$ denotes triangle rasterization. 
And the reliability score is the mask IoU between $S_t$ and $\widehat{S}_t^{\mathrm{vis}}$.
If the pose is unavailable or the object mask is empty, we set $\gamma_t=0$. Given a threshold $\tau_{\mathrm{IoU}}$, the trusted-frame indicator is
\begin{equation}
m_t=\mathbb{1}\!\left[\gamma_t\ge \tau_{\mathrm{IoU}}\right].
\end{equation}
Let $(\mathbf{r}_t^{\mathrm{raw}},\mathbf{u}_t^{\mathrm{raw}})$ denote the raw pose in the diffusion representation, where $\mathbf{r}_t$ is the 6D rotation representation and $\mathbf{u}_t$ is the translation. The filtered 6DoFs is
\begin{equation}
\begin{aligned}
\mathbf{r}_t^{\mathrm{filtered}} &=
\begin{cases}
\mathbf{r}_t^{\mathrm{raw}}, & m_t=1,\\
\mathbf{0}, & m_t=0,
\end{cases} \qquad 
\mathbf{u}_t^{\mathrm{filtered}} =
\begin{cases}
\mathbf{u}_t^{\mathrm{raw}}, & m_t=1,\\
\mathbf{0}, & m_t=0.
\end{cases}
\end{aligned}
\end{equation}

\mypara{Interaction with HOI diffusion.}
The HOI diffusion model receives both the motion hypothesis and the IoU-filtered trusted branch. Let $g_t\in\{L,R,\varnothing\}$ denote the grasp label computed roughly by hand-object distance, where $g_t=\varnothing$ means that no grasp is active. The trusted 6DoF condition $\widetilde{\mathbf{o}}_t^{\mathrm{cond}}$ is provided only when both trust and grasp validity hold, otherwise padded with 0.
During HOI Diffusion sampling, it predicts trajectory only on untrusted grasp frames:
\begin{equation}
\Omega_{\mathrm{pred}}
=
\left\{
t \;\middle|\; g_t\neq\varnothing,\; m_t=0
\right\}.
\end{equation}
Thus, the IoU filter determines which frames act as hard anchors and which are corrected or completed by the HOI prior.

\begin{equation}
\begin{aligned}
\text{SAM3D}
&\;\longrightarrow\; \text{MEMFOF + PnP}
\;\longrightarrow\; \text{interpolation} \\
&\;\longrightarrow\; \text{mask-IoU filtering}
\;\longrightarrow\; \text{HOI diffusion}
\end{aligned}
\end{equation}

The key design is to keep temporal propagation conservative and geometry-driven, and to invoke the generative prior only where the trajectory is marked as untrusted and grasped.

\begin{algorithm}[t]
\caption{SAM3D--MEMFOF Tracking with HOI Diffusion Infilling}
\label{alg:sam3d_memfof_hoi}
\begin{algorithmic}[1]
\REQUIRE $\{I_t,D_t,K_t,S_t,\mathbf{P}_t,H_t,g_t\}_{t=1}^T$, $\tau_{\mathrm{IoU}}$, HOI context $\mathbf{c}_{\mathrm{HOI}}^{1:T}$
\ENSURE Raw poses $\{T_t\}_{t=1}^T$, trusted branch $\{\mathbf{o}_t^{\mathrm{trusted}}\}_{t=1}^T$, completed poses $\{\widehat{\mathbf{o}}_t\}_{t=1}^T$

\STATE Select anchor $a$, support views $\mathcal{V}_a$
\STATE $(\mathbf{M}_{\mathbf{s}},T_a) \leftarrow \mathsf{SAM3D}(\{I_t,S_t,\mathbf{P}_t\})$

\FOR{each sampled edge $(t,t')$ in the forward/backward pass from $a$}
    \STATE $F_{t\rightarrow t'} \leftarrow \mathsf{MEMFOF}(t\rightarrow t')$
    \STATE $\mathcal{C}_{t\rightarrow t'} \leftarrow \mathsf{FlowCorr}(\mathbf{M}_{\mathbf{s}},T_t,K_t,S_t,F_{t\rightarrow t'},S_{t'},H_{t'})$
    \STATE $T_{t'} \leftarrow \mathsf{PnP}(\mathcal{C}_{t\rightarrow t'},K_{t'})$
\ENDFOR

\FOR{$t=1$ to $T$}
    \STATE $\widehat{S}_t^{\mathrm{vis}} \leftarrow \mathsf{RenderMask}(\mathbf{M}_{\mathbf{s}},T_t,K_t)\setminus H_t$
    \STATE $\gamma_t \leftarrow \mathsf{IoU}(S_t,\widehat{S}_t^{\mathrm{vis}}),\quad m_t \leftarrow \mathbb{1}[\gamma_t \ge \tau_{\mathrm{IoU}}]$
    \STATE $ \mathbf{o}_t^{\mathrm{filtered}} \leftarrow m_t\,\mathbf{o}_t^{\mathrm{raw}}, \quad \widetilde{\mathbf{o}}_t^{\mathrm{cond}} \leftarrow \mathbb{1}[m_t=1 \wedge g_t\neq\varnothing]\,\mathbf{o}_t^{\mathrm{filtered}}$
\ENDFOR

\STATE $\Omega_{\mathrm{pred}} \leftarrow \{\, t \mid g_t \neq \varnothing,\; m_t = 0 \,\}$
\STATE $\{\widehat{\mathbf{o}}_t\}_{t=1}^T \leftarrow \mathsf{HOIDiffusion}(\{\mathbf{o}_t^{\mathrm{filtered}}\}_{t=1}^T, \{\widetilde{\mathbf{o}}_t^{\mathrm{cond}}\}_{t=1}^T,\mathbf{c}_{\mathrm{HOI}}^{1:T},\Omega_{\mathrm{pred}})$
\RETURN $\{T_t\}_{t=1}^T$, $\{\mathbf{o}_t^{\mathrm{filtered}}\}_{t=1}^T$, $\{\widehat{\mathbf{o}}_t\}_{t=1}^T$
\end{algorithmic}
\end{algorithm}

\mypara{Summary.}
\cref{alg:sam3d_memfof_hoi} summarizes the full Step-3 pipeline.
It first reconstructs a canonical mesh and anchor pose by SAM3D and applies the predicted mesh scale.
It then propagates poses over sampled frames using MEMFOF-induced correspondences and RANSAC-PnP.
Next, the posed mesh is rasterized in each frame to compute a hand-aware mask IoU reliability score, which determines the trusted-frame indicator.
The raw pose is converted to the diffusion representation, from which the filtered 6DoFs and the grasp-gated conditioning signal are constructed.
Finally, HOI diffusion predicts only on untrusted grasp frames, while trusted frames remain fixed as anchors.

Here, 
$\mathsf{FlowCorr}(\cdot)$ denotes flow-based 2D--3D correspondence construction,
$\mathsf{RenderMask}(\cdot)$ denotes mesh rasterization,
$\mathsf{IoU}(\cdot)$ denotes reliability computation,
$\mathsf{HOIDiffusion}(\cdot)$ denotes untrusted 6DoF infilling.

\begin{wrapfigure}{r}{0.46\textwidth} 
    \centering
    \vspace{-15pt}
    \includegraphics[width=\linewidth]{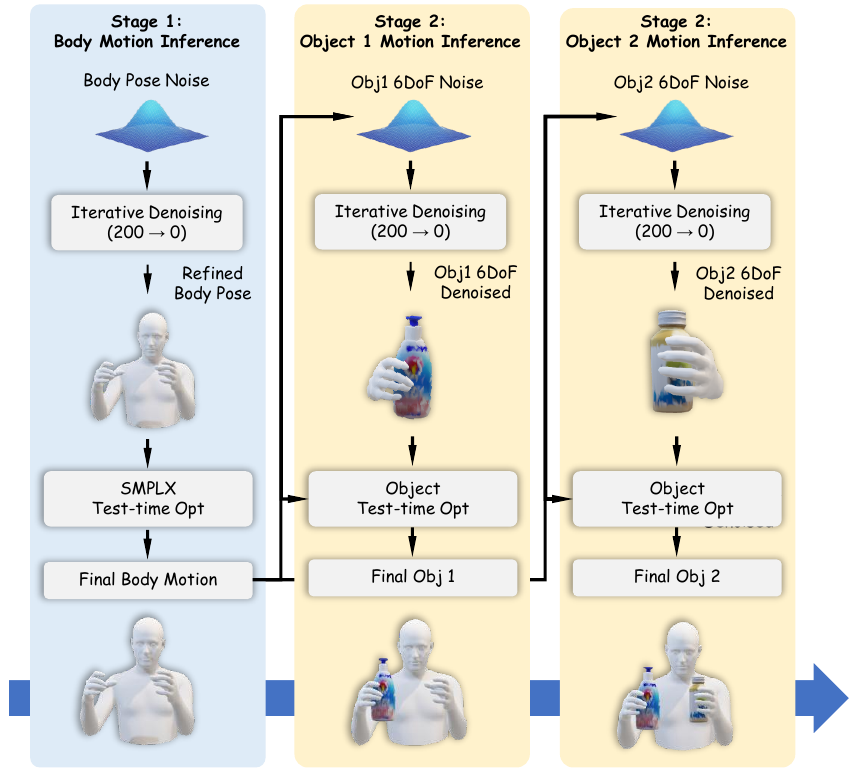}
    \vspace{-10pt}
    \caption{\textbf{Overall algorithm of diffusion inference loop: }(1) Body denoising (2) 6DoF denoising.}
    \label{fig:alg}
    \vspace{-10pt}
\end{wrapfigure}

\section{Multi-Object Inference Loop}

\cref{fig:alg} and \cref{alg:multi_hoi} describe our decoupled multi-object human--object interaction  estimation pipeline. The goal is to recover a temporally coherent upper-body motion together with physically plausible 6DoFs for all interacted objects over a time window of length $T$. The method is decomposed into two stages: the body diffusion model $\mathcal{W}$ first infers human motion, and the HOI diffusion model $\mathcal{H}$ then refines each object trajectory conditioned on the recovered human motion. This decomposition avoids coupling body-pose generation and object-pose generation in a single model while still allowing object inference to benefit from strong human motion priors.

More specifically, the first stage operates on the latent body pose variables $\widehat{\mathbf{z}}_{t_d}^{1:T}$, where $t_d$ denotes the diffusion timestep and $1\!:\!T$ denotes the full temporal sequence. Starting from an initial noisy state at the largest diffusion timestep, the algorithm iteratively applies the body diffusion model until reaching the final denoised body estimate.
\begin{equation}
\widehat{\mathbf{z}}_{t_d-1}^{1:T}
\leftarrow
\mathcal{W}\!\left(
\widehat{\mathbf{z}}_{t_d}^{1:T},
\mathbf{c}^{1:T},
t_d
\right),
\end{equation}
 Here, $\mathbf{c}^{1:T}$ denotes the conditioning features extracted from the central pupil frame (CPF) motion, which provide egocentric temporal cues for body reconstruction. Intuitively, this stage produces a globally consistent upper-body motion that explains the observed egomotion while remaining within the learned human motion manifold. After the diffusion sampling loop, a test-time optimization step further refines the body parameters, as described in \cref{section:tto}.

Once the body motion has been estimated, the second stage infers object motion one object at a time. For each object index $j\in\{1,\dots,J\}$, the algorithm starts from an initial noisy object trajectory $\widehat{\mathbf{o}}_{j,t_d}^{1:T}$ and performs iterative denoising conditioned on both object-specific observations and the body motion:
\begin{equation}
\widehat{\mathbf{o}}_{j,t_d-1}^{1:T}
\leftarrow
\mathcal{H}\!\left(
\widehat{\mathbf{o}}_{j,t_d}^{1:T},
\mathbf{m}_{j}^{1:T},
\mathbf{o}_{\mathrm{filtered},j}^{1:T},
t_d
\right).
\end{equation}
Here, $\mathbf{o}_{\mathrm{filtered},j}^{1:T}$ denotes the reliable object 6DoF estimates obtained from preprocessing, and $\mathbf{m}_{j}^{1:T}$ denotes the object condition features, which encode the filtered object observations, coarse object translation, grasp state, and the hand-motion cues derived from the body prediction. Thus, $\mathcal{H}$ does not infer object motion from object detections alone, it completes and regularizes the object trajectory using both partial object observations and the predicted body motion.
{
\begin{algorithm}[t]
\caption{Multi-Object Inference Loop}
\label{alg:multi_hoi}

\begin{algorithmic}[1]
\REQUIRE $\mathcal{W}$, $\mathcal{H}$, $\mathbf{c}^{1:T}$, $\mathbf{m}_{j}^{1:T}$, $\mathbf{o}_{filtered,j}^{1:T}$, $\mathbf{M}_j$, initial states $\widehat{\mathbf{z}}_{t_d}^{1:T}$, $\{\widehat{\mathbf{o}}_{j,t_d}^{1:T}\}_{j=1}^{J}$
\ENSURE Refined SMPLX Poses $\widehat{\mathbf{z}}_{0}^{1:T}$ and 6DoF $\{\widehat{\mathbf{o}}_{j,0}^{1:T}\}_{j=1}^{J}$
\vspace{0.3em}

\FOR{$t_d = 200, 199, \dots, 0$}
    \STATE $\widehat{\mathbf{z}}_{t_d-1}^{1:T} \leftarrow \mathcal{W}(\widehat{\mathbf{z}}_{t_d}^{1:T}, \mathbf{c}^{1:T}, t_d)$
\ENDFOR
\STATE $\widehat{\mathbf{z}}_{t_0}^{1:T} \leftarrow \mathsf{SMPLXTestTimeOpt}(\widehat{\mathbf{z}}_{t_0}^{1:T})$

\FOR{$j = 1$ \TO $J$}
    \FOR{$t_d = 200, 199, \dots, 0$}
        \STATE $\widehat{\mathbf{o}}_{j,t_d-1}^{1:T} \leftarrow \mathcal{H}(\widehat{\mathbf{o}}_{j,t_d}^{1:T}, \widehat{\mathbf{m}}_{j}^{1:T}, \mathbf{o}_{filtered,j}^{1:T}, t_d)$
    \ENDFOR
    
    \STATE $\widehat{\mathbf{o}}_{j,0}^{1:T} \leftarrow \mathsf{ObjectTestTimeOpt}(\widehat{\mathbf{o}}_{j,0}^{1:T}, \widehat{\mathbf{z}}_{0}^{1:T}, \mathbf{M}_j)$

\ENDFOR

\RETURN $\widehat{\mathbf{z}}_{0}^{1:T}$, $\{\widehat{\mathbf{o}}_{j,0}^{1:T}\}^J_{j=1}$
\end{algorithmic}
\end{algorithm}
}

A key property of Algorithm~\ref{alg:multi_hoi} is that multi-object interaction is handled by placing the object loop outside the HOI diffusion model. The body sequence is inferred once and then shared across all objects, while each object trajectory is generated independently under its own observations and mesh geometry $\mathbf{M}_j$. This design  preserves a common and temporally consistent human motion explanation for the entire interaction sequence. And it naturally scales to a variable number of objects without retraining a joint multi-object generator. The final outputs are the refined body motion $\widehat{\mathbf{z}}_{0}^{1:T}$ and refined object 6DoFs $\{\widehat{\mathbf{o}}_{j,0}^{1:T}\}_{j=1}^{J}$.

From a probabilistic perspective, the algorithm can be interpreted as a sequential conditional generation process:
\begin{equation}
\begin{aligned}
&p\!\left(
\widehat{\mathbf{z}}_{0}^{1:T},
\{\widehat{\mathbf{o}}_{j,0}^{1:T}\}_{j=1}^{J}
\,\middle|\,
\mathbf{c}^{1:T},
\{\mathbf{m}_{j}^{1:T}\}_{j=1}^{J},
\{\mathbf{o}_{\mathrm{filtered},j}^{1:T}\}_{j=1}^{J}
\right)\\
&\approx
p\!\left(
\widehat{\mathbf{z}}_{0}^{1:T}\mid \mathbf{c}^{1:T}
\right)
\prod_{j=1}^{J}
p\!\left(
\widehat{\mathbf{o}}_{j,0}^{1:T}
\,\middle|\,
\widehat{\mathbf{z}}_{0}^{1:T},
\mathbf{m}_{j}^{1:T},
\mathbf{o}_{\mathrm{filtered},j}^{1:T}
\right).
\end{aligned}
\end{equation}
This factorization highlights the role of the recovered body motion as an intermediate representation: the HOI diffusion   uses body motion as a strong prior to resolve object motion ambiguities. As a result, the method achieves temporally complete and physically plausible reconstruction even when object observations are sparse, noisy, or partially missing.

\section{Test-time Optimization}
\label{section:tto}

\subsection{SMPL-X Test-time Optimization}

Given the body diffusion prediction, we refine upper-body SMPL-X parameters by solving
\begin{equation}
\Theta_{\mathrm{human}}^{\star}
=
\arg\min_{\Theta_{\mathrm{human}}}
\mathcal{L}_{\mathrm{human}},
\end{equation}
where
\begin{equation}
\Theta_{\mathrm{human}}
=
\left\{
\theta_{t}^{L},\,
\theta_{t}^{R},\,
\theta_{t}^{\mathrm{sub}}
\right\}_{t=1}^{T}
\cup
\{\beta\}.
\end{equation}
Here, $\theta_{t}^{L}$ and $\theta_{t}^{R}$ denote the left- and right-hand poses, $\theta_{t}^{\mathrm{sub}}$ denotes the optimized upper-body subset of the SMPL-X pose, and $\beta$ is a sequence-shared shape parameter. The optimized upper-body parameters are inserted back into the full SMPL-X pose, while the remaining body parameters are zeros. Global orientation and translation are computed by CPF and body poses.
The objective is
\begin{equation}
\mathcal{L}_{\mathrm{human}}
=
\lambda_{\mathrm{body}}\mathcal{L}_{\mathrm{body}}
+
\lambda_{\mathrm{pose}}\mathcal{L}_{\mathrm{pose}}
+
\lambda_{\mathrm{kp2d}}\mathcal{L}_{\mathrm{kp2d}}
+
\lambda_{\mathrm{wrist}}\mathcal{L}_{\mathrm{wrist}}.
\end{equation}
For brevity, $\operatorname*{avg}$ denotes averaging over the frames, hands, or joints for which the corresponding supervision is available.

\mypara{(1) Pose anchor loss $\mathcal{L}_{\mathrm{body}}$.}
Let $\hat{\theta}_{t}^{L}$, $\hat{\theta}_{t}^{R}$, and $\hat{\theta}_{t}^{\mathrm{sub}}$ be the body diffusion predictions before refinement. We use
\begin{equation}
\mathcal{L}_{\mathrm{body}}
=
\operatorname*{avg}_{t}
\Bigl(
\|\theta_{t}^{\mathrm{sub}}-\hat{\theta}_{t}^{\mathrm{sub}}\|_{2}^{2}
+
\|\theta_{t}^{L}-\hat{\theta}_{t}^{L}\|_{2}^{2}
+
\|\theta_{t}^{R}-\hat{\theta}_{t}^{R}\|_{2}^{2}
\Bigr).
\end{equation}
This term keeps the refined body configuration close to the body diffusion prediction and prevents excessive drift.

\mypara{(2) Hand pose loss $\mathcal{L}_{\mathrm{pose}}$.}
When WiLoR hand-pose estimates $\tilde{\theta}_{t}^{L}$ and $\tilde{\theta}_{t}^{R}$ are available, we align the refined SMPL-X hand poses to them:
\begin{equation}
\mathcal{L}_{\mathrm{pose}}
=
\operatorname*{avg}_{t}
\|\theta_{t}^{L}-\tilde{\theta}_{t}^{L}\|_{2}^{2}
+
\operatorname*{avg}_{t}
\|\theta_{t}^{R}-\tilde{\theta}_{t}^{R}\|_{2}^{2}.
\end{equation}
This term regularizes the hand articulation toward the WiLoR reconstruction while preserving upper-body kinematic consistency.

\mypara{(3) Hand 2D keypoint loss $\mathcal{L}_{\mathrm{kp2d}}$.}
Let $\mathbf{J}_{t,j}^{\mathrm{SMPLX}}$ and $\mathbf{J}_{t,j}^{\mathrm{WiLoR}}$ denote the SMPL-X and WiLoR hand joints in CPF for joint $j$ at frame $t$, and let $\pi_{K_t}(\cdot)$ denote perspective projection under camera intrinsics $K_t$. Denoting by $\mathcal{V}$ the set of jointly visible hand joints, we define
\begin{equation}
\mathcal{L}_{\mathrm{kp2d}}
=
\operatorname*{avg}_{(t,j)\in\mathcal{V}}
\left\|
\pi_{K_t}\!\left(\mathbf{J}_{t,j}^{\mathrm{SMPLX}}\right)
-
\pi_{K_t}\!\left(\mathbf{J}_{t,j}^{\mathrm{WiLoR}}\right)
\right\|_{2}^{2}.
\end{equation}
This term enforces image-space consistency without requiring the optimized 3D body to match the WiLoR geometry exactly.

\mypara{(4) Wrist loss $\mathcal{L}_{\mathrm{wrist}}$.}
To enforce wrist 6DoF consistency in CPF, we combine positional and rotational constraints:
\begin{equation}
\mathcal{L}_{\mathrm{wrist}}
=
\mathcal{L}_{\mathrm{wrist}}^{\mathrm{pos}}
+
\mathcal{L}_{\mathrm{wrist}}^{\mathrm{rot}}.
\end{equation}
Let $\mathbf{w}_{t}^{h}$ and $\tilde{\mathbf{w}}_{t}^{h}$ denote the SMPL-X and WiLoR wrist locations in CPF for hand $h\in\{L,R\}$. The positional term is
\begin{equation}
\mathcal{L}_{\mathrm{wrist}}^{\mathrm{pos}}
=
\operatorname*{avg}_{t,h}
\left\|
\mathbf{w}_{t}^{h}-\tilde{\mathbf{w}}_{t}^{h}
\right\|_{2}^{2}.
\end{equation}
For orientation, let $R_{t,\,w\rightarrow cpf}^{h}\in SO(3)$ be the wrist-to-CPF rotation induced by the current SMPL-X kinematic chain, and let $\tilde{R}_{t,\,w\rightarrow cpf}^{h}$ be the corresponding rotation derived from WiLoR. 
The rotational term is
\begin{equation}
d_{R}(R_{1},R_{2})
=
\left\|
\log\!\left(R_{1}R_{2}^{\top}\right)
\right\|_{2}, \quad
\mathcal{L}_{\mathrm{wrist}}^{\mathrm{rot}}
=
\operatorname*{avg}_{t,h}
d_{R}\!\left(
R_{t,\,w\rightarrow cpf}^{h},
\tilde{R}_{t,\,w\rightarrow cpf}^{h}
\right).
\end{equation}
Together, these terms encourage physically plausible wrist motion while remaining consistent with the hand reconstruction.

\subsection{Object Test-time Optimization}

At test time, we perform a lightweight, fully differentiable refinement of the per-frame object 6DoF pose in CPF. For frame $t$, let
\begin{equation}
T_{t}^{cpf\rightarrow obj}
=
\left(
R_{t}^{cpf\rightarrow obj},\,
\mathbf{t}_{t}^{cpf\rightarrow obj}
\right)
\end{equation}
denote the object pose, and let its inverse be
\begin{equation}
T_{t}^{obj\rightarrow cpf}
=
\left(
R_{t}^{obj\rightarrow cpf},\,
\mathbf{u}_{t}^{obj\rightarrow cpf}
\right)
=
\left(
T_{t}^{cpf\rightarrow obj}
\right)^{-1}.
\end{equation}
Each object is optimized independently by solving
\begin{equation}
\Theta_{\mathrm{obj}}^{\star}
=
\arg\min_{\Theta_{\mathrm{obj}}}
\mathcal{L}_{\mathrm{object}},
\qquad
\Theta_{\mathrm{obj}}
=
\left\{
T_{t}^{cpf\rightarrow obj}
\right\}_{t=1}^{T}.
\end{equation}

The objective is
\begin{equation}
\begin{aligned}
\mathcal{L}_{\mathrm{object}}
={}&\lambda_{\mathrm{obj}}\mathcal{L}_{\mathrm{obj}}
+\lambda_{\mathrm{contact}}\mathcal{L}_{\mathrm{contact}} \\
&+\lambda_{\mathrm{penetration}}\mathcal{L}_{\mathrm{penetration}}
+\lambda_{\mathrm{temporal}}\mathcal{L}_{\mathrm{temporal}} .
\end{aligned}
\end{equation}

with
\begin{equation}
\mathcal{L}_{\mathrm{contact}}
=
\mathcal{L}_{\mathrm{contact}}^{\mathrm{prox}}
+
\mathcal{L}_{\mathrm{contact}}^{\mathrm{noslip}}.
\end{equation}
Thus, consistent with the main paper, the contact term combines a surface-proximity term and a no-slip term. Again, $\operatorname*{avg}$ denotes averaging over the indices for which the corresponding quantities are defined.

To measure object-surface proximity, let $\mathcal{V}_{\mathcal{O}}$ be the object mesh vertices in the object local frame and define the nearest-neighbor distance proxy
\begin{equation}
\tilde{d}_{\mathcal{O}}(\mathbf{x})
=
\min_{\mathbf{v}\in\mathcal{V}_{\mathcal{O}}}
\|\mathbf{x}-\mathbf{v}\|_{2}.
\end{equation}

\mypara{(1) Object anchor loss $\mathcal{L}_{\mathrm{obj}}$.}
Let
\begin{equation}
T_{t}^{obj\rightarrow cpf,(0)}
=
\left(
R_{t}^{obj\rightarrow cpf,(0)},\,
\mathbf{u}_{t}^{obj\rightarrow cpf,(0)}
\right)
\end{equation}
be the HOI Diffusion prediction before refinement. We regularize the refined object pose toward this initialization:
\begin{equation}
\begin{aligned}
\mathcal{L}_{\mathrm{obj}}
=
\operatorname*{avg}_{t}
\Bigl(
& d_{R}\!\left(
R_{t}^{obj\rightarrow cpf},
R_{t}^{obj\rightarrow cpf,(0)}
\right)^{2}
\!\!\!+\!
\left\|
\mathbf{u}_{t}^{obj\rightarrow cpf}
-
\mathbf{u}_{t}^{obj\rightarrow cpf,(0)}
\right\|_{2}^{2}
\Bigr).
\end{aligned}
\end{equation}
This term prevents excessive drift from the HOI Diffusion prediction.

\mypara{(2) Contact loss $\mathcal{L}_{\mathrm{contact}}$.}

\emph{Surface proximity.}
Let $\mathcal{P}_{t}$ be the set of palm sample points from the grasping hand at frame $t$. The proximity term encourages these samples to stay close to a target distance $d_{0}$ from the object surface:
\begin{equation}
\mathcal{L}_{\mathrm{contact}}^{\mathrm{prox}}
=
\operatorname*{avg}_{t,\,\mathbf{p}\in\mathcal{P}_{t}}
\left[
\tilde{d}_{\mathcal{O}}\!\left(
R_{t}^{cpf\rightarrow obj}\mathbf{p}
+
\mathbf{t}_{t}^{cpf\rightarrow obj}
\right)
-
d_{0}
\right]_{+}^{2},
\end{equation}
where $[x]_{+}=\max(x,0)$.

\emph{No-slip contact consistency.}
Let $\mathbf{h}_{t,i}^{cpf}$ be the $i$-th vertex of the grasping hand in CPF, and let
\begin{equation}
\mathbf{h}_{t,i}^{obj}
=
R_{t}^{cpf\rightarrow obj}\mathbf{h}_{t,i}^{cpf}
+
\mathbf{t}_{t}^{cpf\rightarrow obj}
\end{equation}
be the same vertex expressed in the object local frame. For a temporal window $k=1,\dots,K_{c}$, we consider neighboring frames with the same grasping hand and penalize the motion of these hand vertices in the object local frame:
\begin{equation}
\mathcal{L}_{\mathrm{contact}}^{\mathrm{noslip}}
=
\sum_{k=1}^{K_{c}}
\operatorname*{avg}_{(t,t+k),\,i}
w_{t,i}^{(k)}
\left\|
\frac{
\mathbf{h}_{t+k,i}^{obj}
-
\mathbf{h}_{t,i}^{obj}
}{k}
\right\|_{2}^{2},
\end{equation}
where $w_{t,i}^{(k)}\in[0,1]$ is a soft contact weight that increases as the vertex gets closer to the object surface. This term directly encodes the near-zero relative motion assumption at contact and suppresses unnatural sliding.

\mypara{(3) Penetration loss $\mathcal{L}_{\mathrm{penetration}}$.}
Let $\mathcal{B}$ denote sampled points from both hands. We penalize points that violate a safety margin $\delta$ to the object surface:
\begin{equation}
\mathcal{L}_{\mathrm{penetration}}
=
\operatorname*{avg}_{t,\,\mathbf{q}\in\mathcal{B}}
\left[
\delta
-
\tilde{d}_{\mathcal{O}}\!\left(
R_{t}^{cpf\rightarrow obj}\mathbf{q}
+
\mathbf{t}_{t}^{cpf\rightarrow obj}
\right)
\right]_{+}^{2}.
\end{equation}
This term discourages non-physical interpenetration.

\mypara{(4) Temporal loss $\mathcal{L}_{\mathrm{temporal}}$.}
To encourage smooth object motion, we regularize both velocity and acceleration of the inverse pose $T_{t}^{obj\rightarrow cpf}$. For $k=1,\dots,K_{s}$, we define
\begin{equation}
\begin{aligned}
\boldsymbol{\omega}_{t}^{(k)}
&=
\frac{1}{k}
\log\!\left(
R_{t+k}^{obj\rightarrow cpf}
\left(R_{t}^{obj\rightarrow cpf}\right)^{\top}
\right),
&
\mathbf{v}_{t}^{(k)}
&=
\frac{
\mathbf{u}_{t+k}^{obj\rightarrow cpf}
-
\mathbf{u}_{t}^{obj\rightarrow cpf}
}{k},
\\[2mm]
\dot{\boldsymbol{\omega}}_{t}^{(k)}
&=
\frac{
\boldsymbol{\omega}_{t+k}^{(k)}
-
\boldsymbol{\omega}_{t}^{(k)}
}{k},
&
\mathbf{a}_{t}^{(k)}
&=
\frac{
\mathbf{v}_{t+k}^{(k)}
-
\mathbf{v}_{t}^{(k)}
}{k}.
\end{aligned}
\end{equation}
The temporal loss is then
\begin{equation}
\begin{aligned}
\mathcal{L}_{\mathrm{temporal}}
=
\sum_{k=1}^{K_{s}}
\Bigl(
&\alpha_{v}\,
\operatorname*{avg}_{t}
\bigl(
\|\boldsymbol{\omega}_{t}^{(k)}\|_{2}^{2}
+
\|\mathbf{v}_{t}^{(k)}\|_{2}^{2}
\bigr)
\\
&+
\alpha_{a}\,
\operatorname*{avg}_{t}
\bigl(
\|\dot{\boldsymbol{\omega}}_{t}^{(k)}\|_{2}^{2}
+
\|\mathbf{a}_{t}^{(k)}\|_{2}^{2}
\bigr)
\Bigr).
\end{aligned}
\end{equation}
This term suppresses unstable rotational and translational changes and improves trajectory smoothness.

\end{document}